\documentclass[journal,twocolumn]{IEEEtran}

\usepackage{graphicx,times,amsmath,amssymb,cite,subfigure,stfloats,booktabs, url,multirow,afterpage}

\usepackage{graphicx}

\usepackage{array}
\usepackage{booktabs}

\usepackage{commath,amsfonts,algorithm,algorithmic}
\usepackage{cite,multirow,bm,multicol,booktabs,threeparttable,extpfeil}

\allowdisplaybreaks

\begin{document}
\title{Spatial Distillation based Distribution Alignment (SDDA) for Cross-Headset EEG Classification}

\author{Dingkun~Liu, Siyang~Li, Ziwei Wang, Wei Li, and Dongrui~Wu,~\IEEEmembership{Fellow,~IEEE}
\thanks{D.~Liu, S.~Li, Z.~Wang, W.~Li and D.~Wu are with the Ministry of Education Key Laboratory of Image Processing and Intelligent Control, School of Artificial Intelligence and Automation, Huazhong University of Science and Technology, Wuhan 430074, China. They are also with the Shenzhen Huazhong University of Science and Technology Research Institute, Shenzhen, 518000 China.}
\thanks{This research was supported by Shenzhen Science and Technology Program JCYJ20220818103602004.}
\thanks{Corresponding Authors: Wei Li (liwei0828@mail.hust.edu.cn) and Dongrui Wu (drwu09@gmail.com).}}

\maketitle

\begin{abstract}
A non-invasive brain-computer interface (BCI) enables direct interaction between the user and external devices, typically via electroencephalogram (EEG) signals. However, decoding EEG signals across different headsets remains a significant challenge due to differences in the number and locations of the electrodes. To address this challenge, we propose a spatial distillation based distribution alignment (SDDA) approach for heterogeneous cross-headset transfer in non-invasive BCIs. SDDA uses first spatial distillation to make use of the full set of electrodes, and then input/feature/output space distribution alignments to cope with the significant differences between the source and target domains. To our knowledge, this is the first work to use knowledge distillation in cross-headset transfers. Extensive experiments on six EEG datasets from two BCI paradigms demonstrated that SDDA achieved superior performance in both offline unsupervised domain adaptation and online supervised domain adaptation scenarios, consistently outperforming 10 classical and state-of-the-art transfer learning algorithms.
\end{abstract}


\section{Introduction}

A brain-computer interface (BCI) serves as a direct communication pathway between the human or animal brain and an external device~\cite{nicolas2012brain}. There are generally three types of BCIs: Invasive, non-invasive, and semi-invasive. This paper focuses on electroencephalogram (EEG) based non-invasive BCIs.

\begin{figure}[htbp]         \centering
\includegraphics[width=\linewidth,clip]{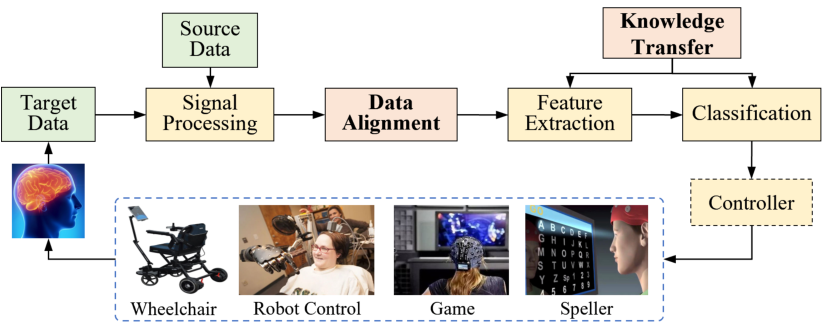}
\caption{Transfer learning for BCIs.} \label{fig:closed_loop}
\end{figure}

Despite the advantages of cost effectiveness and convenience, EEGs suffer from substantial individual differences and non-stationarity. Transfer learning has been extensively studied in the literature to address individual differences, enabling the transfer of data/knowledge from source domains to facilitate calibration in the target domain~\cite{Wu2022}. Fig.~\ref{fig:closed_loop} depicts the flowchart of transfer learning for BCIs.

Most existing transfer learning approaches focus on cross-subject or cross-session transfers using an identical input space~\cite{lotte2018review}, which are not readily applicable to cross-headset transfers, where disparities in the number and locations of EEG electrodes between the source and target headsets result in non-identical input spaces. For cross-headset transfer, a typical strategy is to crop EEG signals with more channels to match those with fewer channels, causing substantial spatial information loss and hence suboptimal transfer performance.

This paper considers heterogeneous transfer learning, extending beyond traditional and simpler homogeneous approaches. Theoretically, transfer learning considers four discrepancies between the source and target domains: 1) marginal probability distribution; 2) conditional probability distribution; 3) input (feature) space; and, 4) output (label) space. Homogeneous transfer learning focuses on aligning the marginal and conditional probability distributions, under the assumption that different domains share an identical input space. In contrast, heterogeneous transfer learning, as considered in this paper, additionally accounts for discrepancies in the input space between the source and target domains.

We propose spatial distillation based distribution alignment (SDDA) for cross-headset heterogeneous transfer learning. To the best of our knowledge, this is the first work to handle the input space discrepancies for cross-headset transfer, by utilizing information from extra channels in the labeled source dataset through knowledge distillation.

Our main contributions are:
\begin{enumerate}
\item  We propose spatial distillation (SD) for heterogeneous transfer learning among different EEG headsets, leveraging knowledge from EEG signals with more channels to improve those with fewer channels. This approach effectively addresses the challenge of limited spatial information utilization inherent in fewer-channel headsets.
\item  We introduce a distribution alignment (DA) strategy that aligns the source and target domains comprehensively in multiple stages of the model, i.e., input/feature/output spaces. Unlike previous approaches that rely on single-stage alignment, the proposed DA more effectively bridges the domain gaps, ensuring robust transfer.
\item  Extensive experiments on multiple EEG datasets, covering both motor imagery (MI) and P300 paradigms, validated the superior performance of SDDA, which consistently outperformed state-of-the-art homogeneous transfer learning approaches in both offline and online calibration scenarios.
\end{enumerate}

The remainder of this paper is organized as follows: Section~\ref{sect:related} introduces related work. Section~\ref{sect:approach} proposes SDDA. Section~\ref{sect:exp} presents the experiment results. Finally, Section~\ref{sect:conclusion} draws conclusions.

\section{Related Work} \label{sect:related}

This section introduces related works on transfer learning and cross-headset transfer in EEG-based BCIs.

\subsection{Transfer Learning} \label{sect:domaintl}

Transfer learning utilizes data/knowledge in one or more source domains to enhance the analysis in a target domain. By minimizing discrepancies between the source and target data distributions, a classifier built on the source data can perform well on unknown target data~\cite{wu2020transfer}.

Various approaches have been proposed to measure cross-domain discrepancies, including maximum mean discrepancy (MMD)~\cite{gretton2012kernel}, higher-order statistical metrics~\cite{chen2020homm}, the optimized transportation distance~\cite{courty2017joint}, etc. Long \textit{et al.}~\cite{long2015learning} adapted MMD with multiple kernels to capture more comprehensive data statistics. Instead of direct calculation, Ganin \textit{et al.}~\cite{Ganin2016} introduced domain adversarial neural networks (DANN), which simultaneously optimizes a domain discriminator and a feature extractor to reduce the discrepancies between the source and target domains.

Later approaches additionally leverage category information to minimize distribution shifts. Long \textit{et al.}~\cite{Long2017JAN} proposed joint adaptation networks (JAN), which align the joint distributions by a joint MMD metric that takes class-wise predictions into calculation. They further introduced conditional domain adversarial networks (CDAN)~\cite{long2018conditional}, which includes adversarial learning and entropy minimization. Zhang \emph{et al.}~\cite{zhang2019bridging} proposed margin disparity discrepancy (MDD), a measurement for comparing the distributions with asymmetric margin loss and easier minimax optimization in domain adaptation. Chen \textit{et al.}~\cite{jin2020minimum} proposed minimum class confusion (MCC), which reduces the class confusion based on the target domain predictions. Liang \textit{et al.}~\cite{liang2020we} proposed Source HypOthesis Transfer (SHOT), which minimizes the prediction uncertainty and maximizes the prediction diversity. Li \textit{et al.}~\cite{li2021imbalanced} proposed imbalanced source-free domain adaptation (ISFDA) to address class imbalance and label shifts, utilizing secondary label correction, curriculum sampling, and intra-class tightening with inter-class separation.

\subsection{Cross-Headset Transfer} \label{sect:headsettl}

The above works mainly consider homogeneous domain adaptation; however, the feature spaces of the source and target domains are different in heterogeneous cross-headset transfer.

Recently, a few cross-dataset transfer learning approaches have been explored in EEG-based BCIs. Wu \emph{et al.}~\cite{drwuTNSRE2016} proposed active weighted adaptation regularization, which integrates domain adaptation and active learning, for cross-headset transfer. Xu \textit{et al.}~\cite{xu2021enhancing} combined alignment and adaptive batch normalization in neural networks to improve generalization, integrating also manifold embedded knowledge transfer~\cite{Zhang2020}. Zaremba \textit{et al.}~\cite{zaremba2022cross} performed cross-subject transfer for MI-based BCIs, achieving promising performance in both within-dataset and across-dataset settings. Xie \textit{et al.}~\cite{xie2023cross} proposed a pretraining-based cross-dataset transfer learning approach for MI classification, leveraging hard parameter sharing to improve the accuracy and robustness across MI tasks with minimal fine-tuning. Jin \textit{et al.}~\cite{jin2024cross} proposed a cross-dataset adaptive domain selection framework for MI-based BCIs, combining domain selection, data alignment, and enhanced common spatial patterns (CSP) to improve the classification accuracy while minimizing the calibration time.

All above approaches, except \cite{drwuTNSRE2016}, used only the identical subset of EEG channels in the source and target datasets, simplifying the problem to homogeneous transfer but significantly reducing spatial information utilization.

\section{SDDA} \label{sect:approach}

\begin{figure*}[htbp]\centering
\includegraphics[width=\linewidth,clip]{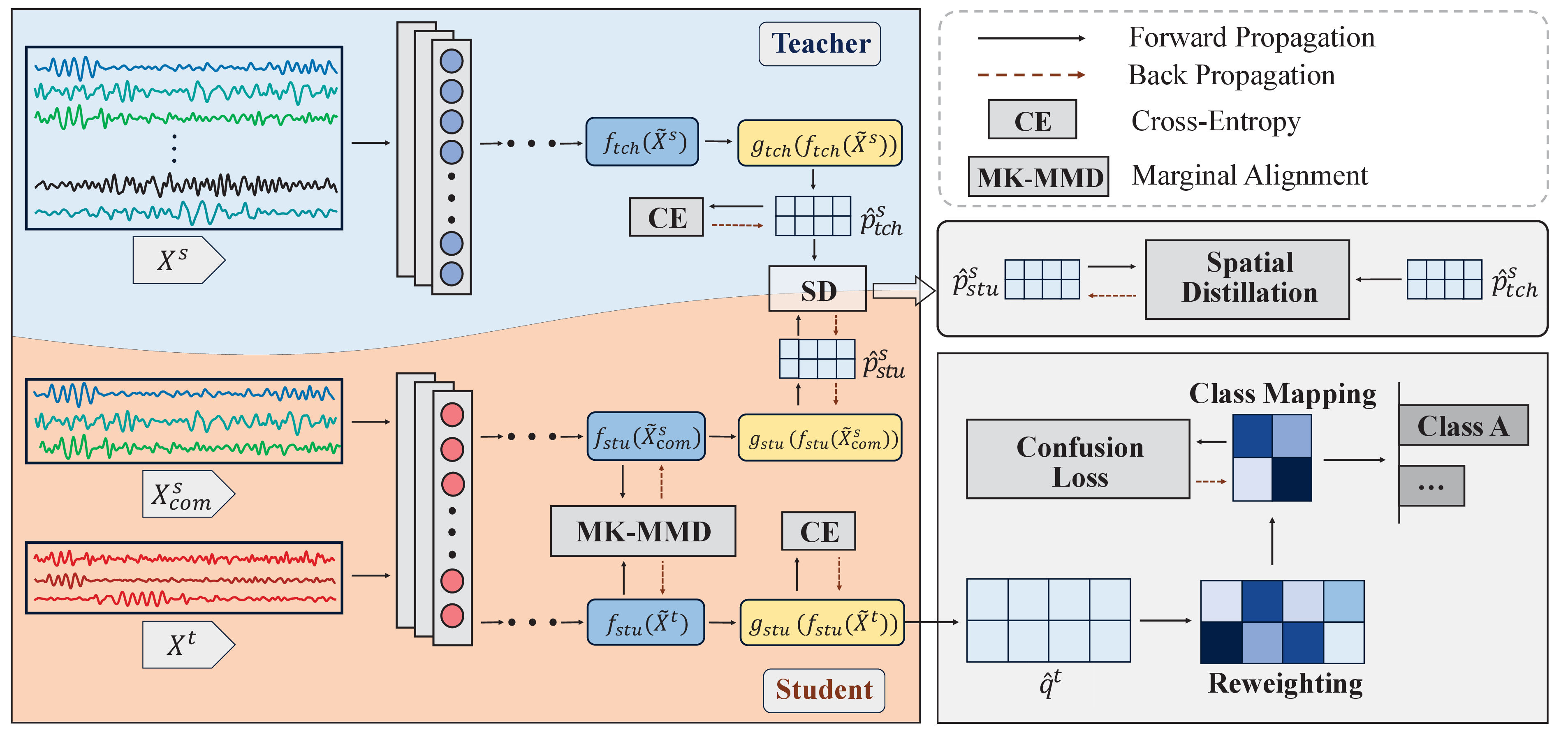}
\caption{Architecture of the proposed SDDA for cross-headset EEG classification. The source data with a full set of electrodes are used to train the teacher model. The student model is trained on source data using common electrodes with the target domain. Target data are incorporated to align the probability distributions and reduce the prediction confusion. Cross-entropy loss is applied to the labeled source data and a small amount of labeled target calibration data.} \label{fig:approach}
\end{figure*}

This section introduces our proposed SDDA for cross-headset EEG classification, as illustrated in Fig.~\ref{fig:approach}. SD enables transfer from a higher dimensional feature space to a lower one, eliminating electrode discrepancies in the spatial domain. DA further mitigates the distribution shift from three different aspects. Table~\ref{notations} summarizes the main notations used throughout this paper.

\begin{table}[h]
\centering
\renewcommand{\arraystretch}{1.4} 
\caption{Notations used in this paper.}
\label{notations}
\scriptsize 
\begin{tabular}{c|>{\centering\arraybackslash}p{0.73\linewidth}}
\toprule
\textbf{Notation} & \textbf{Description} \\
\midrule
$\mathcal{C}$ & The number of classes \\
$\{(X_i^\mathrm{s},y_i^\mathrm{s})\}_{i=1}^{n_\mathrm{s}}$ & $n_\mathrm{s}$ labeled source EEG trials \\
$\{X_i^\mathrm{t}\}_{i=1}^{n_\mathrm{t}}$ & $n_\mathrm{t}$ unlabeled target EEG trials in UDA \\
$\{(X_i^\mathrm{t},y_i^\mathrm{t})\}_{i=1}^{n_l}$ & $n_l$ labeled target EEG trials in SDA  \\
$f_{\mathrm{tch}}$ & Feature extractor of the teacher model \\
$g_{\mathrm{tch}}$ & Classifier of the teacher model \\
$f_{\mathrm{stu}}$ & Feature extractor of the student model \\
$g_{\mathrm{stu}}$ & Classifier of the student model \\
$\widetilde{X}_i^\mathrm{s}$ & The $i$-th aligned EEG trial in the source domain \\
$\widetilde{X}_i^\mathrm{t}$ & The $i$-the aligned EEG trial in the target domain \\
$k$ & The convex combination of $m$ individual kernels \\
$\hat{p}_\mathrm{tch}^\mathrm{s}$ & Teacher model logit prediction for the source EEG
 trials\\
$\hat{p}_\mathrm{stu}^\mathrm{s}$ & Student model logit prediction for the cropped source EEG trials\\
$q_{ij}$ & Logit prediction for the $i$-th target EEG trial and $j$-th category \\
$v_i$ & Uncertainty weight for the $i$-th target trial \\
$J(\cdot, \cdot)$ & Cross-entropy classification loss \\
\bottomrule
\end{tabular}
\end{table}

\subsection{Problem Definition}

Given $n_\mathrm{s}$ labeled source trials $\{(X_i^\mathrm{s},y_i^\mathrm{s})\}_{i=1}^{n_\mathrm{s}}$, where $X_i^\mathrm{s} \in  \mathbb{R}^{C_\mathrm{s} \times T}$ and $y_i^\mathrm{s} \in  \{ 1, 2, ..., \mathcal{C} \}$ ($\mathcal{C}$ is the number of classes), and $n_\mathrm{t}$ unlabeled target trials $\{X_i^\mathrm{t}\}_{i=1}^{n_\mathrm{t}}$, where $X_i^\mathrm{t}\in \mathbb{R}^{C_\mathrm{t} \times T}$ and $C_\mathrm{t} \le C_\mathrm{s}$ (the target domain electrodes are a subset of those in the source domain), the goal is to learn a model that accurately predicts the target trial labels $\{y_i^\mathrm{t}\}_{i=1}^{n_\mathrm{t}}$.

We consider two scenarios:
\begin{enumerate}
\item Online supervised domain adaptation (SDA), where $n_l$ ($n_l \ll n_\mathrm{s}$) labeled target trials $\{(X_i^\mathrm{t},y_i^\mathrm{t})\}_{i=1}^{n_l}$ are available, and the target test trials $\{X_i^\mathrm{t}\}_{i=1}^{n_\mathrm{t}}$ are inaccessible during training.
\item Offline unsupervised domain adaptation (UDA), where no labeled data are available from the target domain, but the unlabeled target test trials $\{X_i^\mathrm{t}\}_{i=1}^{n_\mathrm{t}}$ are accessible during training.
\end{enumerate}

\subsection{Spatial Distillation}

Traditional generalization error bounds are typically derived under the assumption that the source and target domains share an identical feature space~\cite{long2015learning}, enabling aligning distributions by the same model architecture. However, in the challenging heterogeneous scenario, source and target data are collected from different EEG headsets with varying number/locations of electrodes, rendering traditional theoretical results inapplicable to heterogeneous settings.

A novel SD approach is proposed here to address this challenge. A teacher model $g_{\mathrm{tch}} \circ f_{\mathrm{tch}}$, trained on the full set of electrodes in the source domain, transfers its knowledge to the student model $g_{\mathrm{stu}} \circ f_{\mathrm{stu}}$, which uses only the common subset of channels between the two domains. Note here $f_{\mathrm{tch}}$ and $f_{\mathrm{stu}}$ represent the feature extractors for the teacher and student models, respectively, and $g_{\mathrm{tch}}$ and $g_{\mathrm{stu}}$ the corresponding classifiers. SD facilitates semantic alignment between the teacher and student models by minimizing their output distribution discrepancies, ensuring that the student model, despite using a reduced set of EEG channels, closely approximates the output of the teacher model trained on the full set of electrodes.

More specifically, the distillation loss $L_{SD}$ is computed as:
\begin{align}
L_\mathrm{SD}& = {\mathcal{T}}^2 \cdot D_{\mathrm{KL}}(\hat{p}_\mathrm{stu}^\mathrm{s} || \hat{p}_\mathrm{tch}^\mathrm{s})\nonumber\\
&={\mathcal{T}}^2 \cdot\sum_{i=1}^{\mathcal{C}} \hat{p}_\mathrm{stu}^\mathrm{s}(i) \log \frac{\hat{p}_\mathrm{stu}^\mathrm{s}(i)}{\hat{p}_\mathrm{tch}^\mathrm{s}(i)}, \label{eq:SD}
\end{align}
where $\mathcal{T}$ is the temperature, $D_\mathrm{KL}$ is the Kullback-Leibler divergence between two probability distributions over $\mathcal{C}$ categories. $\hat{p}_\mathrm{tch}^\mathrm{s}$ and $\hat{p}_\mathrm{stu}^\mathrm{s}$ represent the prediction probabilities of the teacher and student models, respectively. Note that the teacher model is trained on source domain data with all available channels, whereas the student model is trained on the same EEG trials but only a common subset of channels with the target domain.

SD facilitates the transfer of information from the full set of electrodes to the reduced subset, allowing both the teacher and student models to jointly learn high-level semantic features from distinct feature spaces in the source domain. SD maximizes the spatial feature utilization of EEG signals and implicitly mitigates the discrepancies between the source and target domains, enabling effective transfer across heterogeneous EEG headsets.

\subsection{Distribution Alignment}

While SD achieves feature space alignment, significant disparities in the probability distributions between the source and target domains after transformation remain a critical challenge for constructing an effective classifier. To address this, we introduce DA, which further reduces the distribution shifts via:
\begin{enumerate}
\item Input-space data normalization using session-wise Euclidean alignment (EA)~\cite{he2019transfer}.
\item Feature-space marginal distribution matching using MMD~\cite{gretton2012kernel}.
\item Output-space uncertainty minimization using the confusion loss~\cite{jin2020minimum}.
\end{enumerate}

\subsubsection{Session-wise EA}

EEG data are inherently non-stationary. Data normalization, often referred to as whitening, is a commonly employed preprocessing technique in machine learning to suppress noise. It not only helps mitigate marginal distribution shifts between the source and target domains, but also enhances the consistency within the source domain, particularly when EEG data are collected from multiple subjects.

Assume a session has $n$ EEG trials $\{X_i\}_{i=1}^n$. EA first computes the mean covariance matrix of all trials:
\begin{align}
\bar{R}=\frac{1}{n}\sum_{i=1}^n X_iX_i^\top, \label{eq:EA-Ref}
\end{align}
and then performs the transformation:
\begin{align}
\widetilde{X}_i = \bar{R}^{-1/2} X_i. \label{eq:s-EA}
\end{align}

The mean covariance matrix of $\{\widetilde{X}_i\}_{i=1}^n$ becomes an identity matrix, i.e., the discrepancy in second-order statistics are reduced. $\{\widetilde{X}_i\}_{i=1}^n$ are then used to replace the original trials $\{X_i\}_{i=1}^n$ in all subsequent calculations.

\subsubsection{Marginal Alignment (MA)}

EA aligns the input EEG data, whereas covariate shift can still happen after feature extraction. Multi-kernel MMD (MK-MMD)~\cite{long2015learning} is used to further reduce the substantial marginal distribution differences in the feature space (also called deep representation space in deep learning) between the source and target domains. MK-MMD minimizes the discrepancy between the source and target domains by aligning their feature distributions in multiple latent feature spaces, providing a more flexible and precise measure of domain divergence than a single kernel.

Let $\mathcal{K}$ be a combination of $m$ individual kernels $\mathcal{K}_i$:
\begin{align}
\mathcal{K}=\sum_{i=1}^m\beta_i \mathcal{K}_i, \quad \mathrm{s.t.} \quad\sum_{i=1}^m\beta_i=1 \text{ and } \beta_i\geq0, \forall i, \label{eq:MK}
\end{align}
where $\{\beta_i\}_{i=1}^m$ are the non-negative kernel weights. The marginal alignment loss function is then:
\begin{align}
L_\mathrm{MA} = \Big\| \mathbb{E} \left[ \phi(f_{\mathrm{stu}}(\widetilde{X}_\mathrm{com}^\mathrm{s})) \right] - \mathbb{E} \left[ \phi(f_{\mathrm{stu}}(\widetilde{X}^\mathrm{t})) \right] \Big\|_{\mathcal{H}}^2, \label{eq:MA}
\end{align}
where $\widetilde{X}_\mathrm{com}^\mathrm{s}$ and $\widetilde{X}^\mathrm{t}$ represent the aligned source EEG data and the target EEG data with the common channels after EA, respectively. $L_\mathrm{MA}$ is the squared MK-MMD discrepancy computed in the reproducing kernel Hilbert space (RKHS) $\mathcal{H}$, where $\mathbb{E}[\cdot]$ represents the mean embedding and $\phi(\cdot)$ denotes the feature mapping in the RKHS induced by the kernel $\mathcal{K}$. Specifically, $\mathcal{K}\left(f_{\mathrm{stu}}(\widetilde{X}_\mathrm{com}^\mathrm{s}), f_{\mathrm{stu}}(\widetilde{X}^\mathrm{t})\right) = \left\langle \phi(f_{\mathrm{stu}}(\widetilde{X}_\mathrm{com}^\mathrm{s})), \phi(f_{\mathrm{stu}}(\widetilde{X}^\mathrm{t})) \right\rangle_{\mathcal{H}}$, where $\langle \cdot, \cdot \rangle_{\mathcal{H}}$ denotes the inner product in the RKHS $\mathcal{H}$. By minimizing $L_\mathrm{MA}$, the marginal alignment loss reduces the discrepancy between the source and target distributions in the RKHS, facilitating the model to learn domain-invariant feature representations.

The marginal alignment loss is utilized to optimize the student model, guiding it to learn representations that are shared across the source and target domains.

\subsubsection{Confusion Loss (CL)}

CL~\cite{jin2020minimum} is used to further reduce class-level discrepancies, by reducing the prediction uncertainty in the target domain.

To achieve this,  the prediction uncertainty weight induced by entropy for each trial is computed:
\begin{align}
v_i=1+\exp\left(\sum_{j=1}^\mathcal{C}\hat{q}_{ij}\log\hat{q}_{ij}\right), \label{eq:reweight}
\end{align}
where $\mathcal{C}$ is the number of categories, and $\hat{q}_{ij}$ is the softened logit to reduce the overconfidence of the predictions~\cite{Guo2017}:
\begin{align}
\hat{q}_{ij}=\frac{\exp\!\left(\frac{q_{ij}}{\tau}\right)}{\sum_{j'=1}^\mathcal{C}\exp\!\left(\frac{q_{ij'}}{\tau}\right)}, \label{eq:temp}
\end{align}
in which $q_{ij}$ is the logit (the outputs of the classifier $g$ before converted into probabilities by softmax) of the $i$-th target trial being classified into the $j$-th category, and $\tau$ is the temperature.

CL is then computed as:
\begin{align}
L_\mathrm{CL}=\left(\sum_{j=1}^\mathcal{C}\sum_{j'=1}^\mathcal{C}l_{jj'}-\sum_{j=1}^\mathcal{C}l_{jj}\right)/\mathcal{C}, \label{eq:L_conf}
\end{align}
where
\begin{align}
l_{jj^{\prime}}=\sum_{i=1}^{n} q_{ij}v_iq_{ij^{\prime}}
\label{eq:confusion}
\end{align}
denotes the contribution of the interaction between the $j$-th and $j^{\prime}$-th categories in the model predictions. Here, $n$ is the number of EEG trails, i.e., $n=n_l$ in SDA and $n=n_\mathrm{t}$ in UDA.

Essentially, $L_\mathrm{CL}$ measures the discrepancy between off-diagonal elements (indicating inter-class confusion) and diagonal elements (representing correct classifications), reducing class confusion and enhancing generalization to the target domain.

\subsection{Summary}

Let $\widetilde{X}^\mathrm{s}$ be the source EEG data after EA, with full set of source domain channels. As before, let $\widetilde{X}_\mathrm{com}^\mathrm{s}$ be aligned source EEG data after EA, with only the common channels of the two domains; and, $\widetilde{X}^\mathrm{t}$ be the target EEG data after EA. The teacher model is trained on $\widetilde{X}^\mathrm{s}$, using loss function:
\begin{align}
L^\mathrm{UDA}_\mathrm{tch}=\frac1{n_\mathrm{s}}\sum_{i=1}^{n_\mathrm{s}}J\left(g_{\mathrm{tch}}(f_{\mathrm{tch}}(\widetilde{X}_i^\mathrm{s})),y_i^\mathrm{s}\right), \label{eq:tch}
\end{align}
where $J(\cdot, \cdot)$ is the cross-entropy loss.

The student model is trained on both $X_\mathrm{com}^\mathrm{s}$ and $X^\mathrm{t}$. In the offline UDA scenario, the loss function is:
\begin{align}
\begin{aligned}
L^\mathrm{UDA}_\mathrm{stu} &=\frac{1}{n_s}\sum_{i=1}^{n_s}J\left( g_{\mathrm{stu}}(f_{\mathrm{stu}}
(\widetilde{X}_{\mathrm{com},i}^\mathrm{s})),y_i^\mathrm{s}\right)+\alpha L_\mathrm{SD} \\
&+ \beta L_\mathrm{MA} + \gamma L_\mathrm{CL} ,\label{eq:stu-uda}
\end{aligned}
\end{align}
where $\alpha$, $\beta$ and $\gamma$ are trade-off parameters.

In the online SDA scenario, where $n_l$ labeled target data are available, the loss function of the student model is:
\begin{align}
\begin{aligned}
L^\mathrm{SDA}_\mathrm{stu} &= \frac{1}{n_\mathrm{s}}\sum_{i=1}^{n_\mathrm{s}} J\left(g_{\mathrm{stu}}(f_{\mathrm{stu}}(\widetilde{X}_{\mathrm{com},i}^\mathrm{s}), y_i^\mathrm{s}\right) \\
&+ \frac{1}{n_l}\sum_{i=1}^{n_l} J\left(g_{\mathrm{stu}}(f_{\mathrm{stu}}(\widetilde{X}_i^\mathrm{t}), y_i^\mathrm{t}\right) \\
&+ \alpha L_\mathrm{SD} + \beta L_\mathrm{MA} + \gamma L_\mathrm{CL}. \label{eq:stu-sda}
\end{aligned}
\end{align}

In summary, the loss for the student model combines the cross-entropy loss for all available labeled data, and regularization terms for spatial distillation, feature-space alignment, and output-space alignment. The student model is then employed for final inference.

Algorithm~\ref{alg:alg_uda} gives the pseudo-code of SDDA.

\begin{algorithm}[h!]
 \caption{Spatial Distillation based Distribution Alignment (SDDA) for cross-headset transfer.}
 \label{alg:alg_uda}
 \begin{algorithmic}
\REQUIRE Source domain labeled data $\{(X_i^\mathrm{s},y_i^\mathrm{s})\}_{i=1}^{n_\mathrm{s}}$;\\
 Target domain labeled data $\{(X_i^\mathrm{t},y_i^\mathrm{t})\}_{i=1}^{n_l}$ ($n_l \ll n_\mathrm{s}$) (unavailable in offline UDA);\\
 Target domain unlabeled test data $\{X_i^\mathrm{t}\}_{i=1}^{n_\mathrm{t}}$ ;\\
 $g_{\mathrm{tch}} \circ f_{\mathrm{tch}}$, the teacher model;\\
 $g_{\mathrm{stu}} \circ f_{\mathrm{stu}}$, the student model;\\
\ENSURE The classifications $\{\hat{y}_i^\mathrm{t}\}_{i=1}^{n_\mathrm{t}}$ for $\{X_i^\mathrm{t}\}_{i=1}^{n_\mathrm{t}}$.

\STATE //\ \textbf{Step 1: Session-wise EA}
\STATE Perform session-wise EA on $\{(X_i^\mathrm{s},y_i^\mathrm{s})\}_{i=1}^{n_\mathrm{s}}$ by (\ref{eq:EA-Ref}) and (\ref{eq:s-EA}) to obtain $\widetilde{X}^\mathrm{s} = \{\widetilde{X}_i^\mathrm{s}\}_{i=1}^{n_\mathrm{s}}$;
\STATE Perform session-wise EA on $\{(X_i^\mathrm{s},y_i^\mathrm{s})\}_{i=1}^{n_\mathrm{s}}$ using the common channel subset by (\ref{eq:EA-Ref}) and (\ref{eq:s-EA}) to obtain $\widetilde{X}_\mathrm{com}^\mathrm{s} = \{\widetilde{X}_{\mathrm{com}, i}^\mathrm{s}\}_{i=1}^{n_\mathrm{s}}$;
\STATE Perform session-wise EA on $\{(X_i^\mathrm{t},y_i^\mathrm{t})\}_{i=1}^{n_l}$ by (\ref{eq:EA-Ref}) and (\ref{eq:s-EA}) to obtain  $\widetilde{X}^\mathrm{t} = \{\widetilde{X}_i^\mathrm{t}\}_{i=1}^{n_l}$;

\STATE //\ \textbf{Step 2: Feature Extraction}
\STATE Pass $\widetilde{X}^\mathrm{s}$ through $g_{\mathrm{tch}} \circ f_{\mathrm{tch}}$ to get the category logits $\hat{p}_\mathrm{tch}^\mathrm{s}$;
\STATE Pass $\widetilde{X}_\mathrm{com}^\mathrm{s}$ and $\widetilde{X}^\mathrm{t}$ through $f_{\mathrm{stu}}$ to get student model feature representations $f_{\mathrm{stu}}(\widetilde{X}_\mathrm{com}^\mathrm{s})$ and $f_{\mathrm{stu}}(\widetilde{X}^\mathrm{t})$;
\STATE Pass $\widetilde{X}_\mathrm{com}^\mathrm{s}$ and $\widetilde{X}^\mathrm{t}$ through $g_{\mathrm{stu}} \circ f_{\mathrm{stu}}$ to get the category logits $g_{\mathrm{stu}}(f_{\mathrm{stu}}(\widetilde{X}_\mathrm{com}^\mathrm{s})) := \hat{p}_\mathrm{stu}^\mathrm{s}$ and $g_{\mathrm{stu}}(f_{\mathrm{stu}}(\widetilde{X}^\mathrm{t})) := \hat{q}^\mathrm{t}$;

\STATE //\ \textbf{Step 3: Model Training}
\STATE Simultaneously optimize the teacher model $g_{\mathrm{tch}} \circ f_{\mathrm{tch}}$ by minimizing (\ref{eq:tch}), and the student model $g_{\mathrm{stu}} \circ f_{\mathrm{stu}}$ by minimizing (\ref{eq:stu-sda}) in online SDA, or (\ref{eq:stu-uda}) in offline UDA, until convergence;
\STATE //\ \textbf{Step 4: Final Prediction}
\STATE Use the trained student model to obtain predictions of target test trials, $\{\hat{y}_i^\mathrm{t}\}_{i=1}^{n_l}$.
 \end{algorithmic}
 \end{algorithm}

\section{Experiments} \label{sect:exp}

This section performs experiments to validate the effectiveness of SDDA.

\subsection{Datasets}

Two EEG-based BCI paradigms, MI and P300, are considered. MI~\cite{pfurtscheller2001motor} is the cognitive process of imagining the movement of different body parts without actually moving them. Event-related potentials (ERP)~\cite{lees2018review} is the related potential shown in the EEG after the brain responds to a visual, audio, or tactile stimulus. P300, a positive EEG peak occurring approximately 300ms after a rare stimulus, is one of the most frequently used ERPs.

Four MI datasets and two P300 datasets, all from the mother of all BCI benchmark (MOABB)~\cite{Jayaram2018} and summarized in Table~\ref{tab:datasets}, were utilized in the experiments.

\begin{table*}[htpb] \centering
\caption{Summary of the six EEG datasets.}
\label{tab:datasets}
\begin{tabular}{c|c|c|c|c|c|c|c}
\toprule
BCI&\multirow{2}{*}{Dataset} & Number of & Number of & Sampling & Trial Length & Number of Trials  & \multirow{2}{*}{Class Labels} \\
Paradigm& & Subjects & Channels & Rate (Hz) & (seconds) & per Session &  \\
\midrule
\multirow{4}{*}{MI}&BNCI2014001 & 9 & 22 & 250 & 4  & 144 &  left hand, right hand \\
&BNCI2014004 & 9 & 3 & 250 & 4 & 680-760 &  left hand, right hand \\
&BNCI2014002 & 14 & 15 & 512 & 5 & 100 &  right hand, both feet \\
&BNCI2015001 & 12 & 13 & 512 & 5 & 200 &  right hand, both feet \\
\midrule
\multirow{2}{*}{P300}&BNCI2014009 & 10 & 16 & 256 & 0.8 & 576 &  target, non-target \\
&BNCI2014008 & 8 & 8 & 256 & 1 & 4,200 &  target, non-target \\
\bottomrule
\end{tabular}
\end{table*}

\subsection{Experiment Settings}

\begin{figure}[htbp]         \centering
\includegraphics[width=1.0\linewidth,clip]{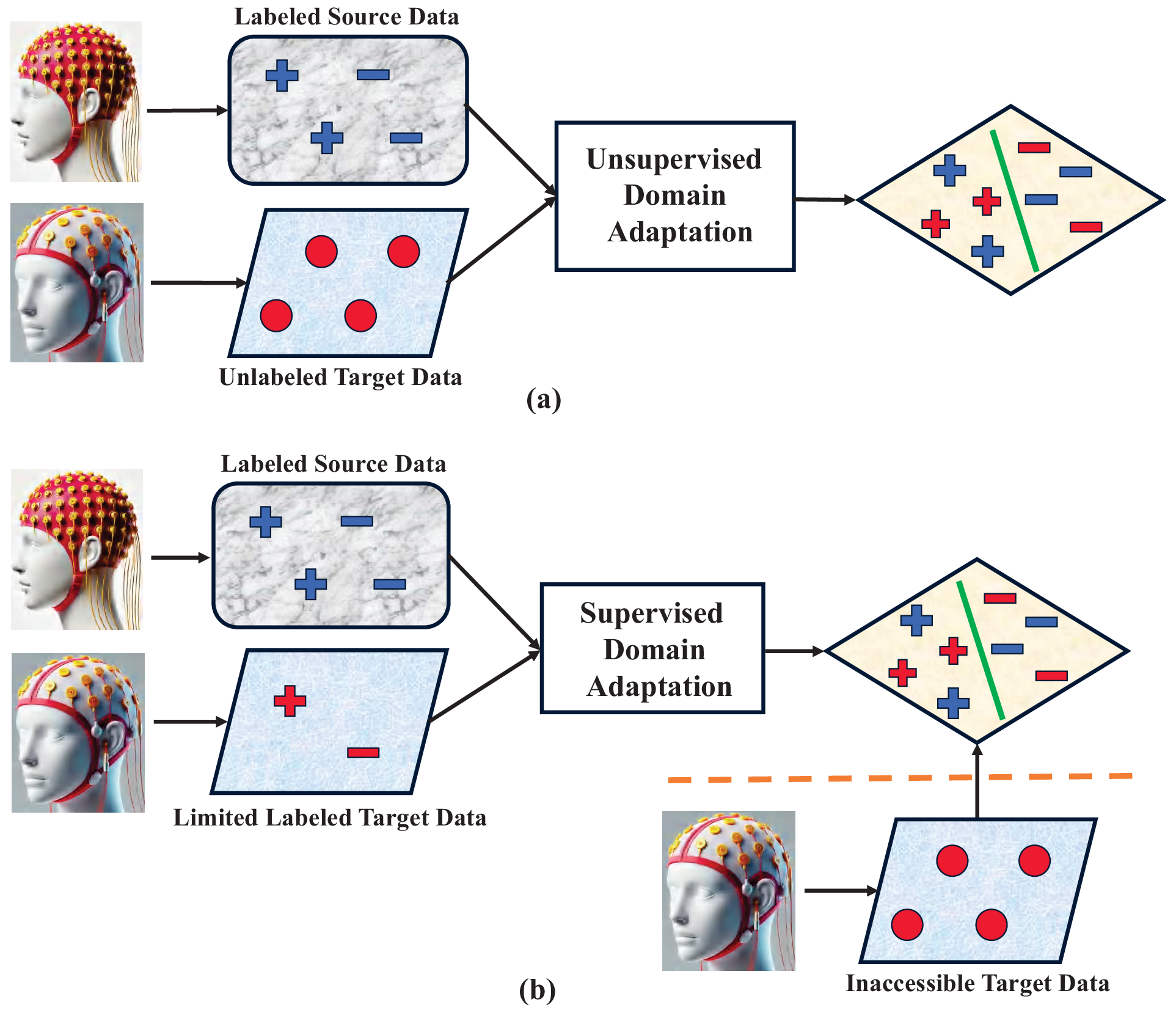}
\caption{Two different cross-headset transfer settings. (a) UDA; and, (b) SDA.} \label{fig:settings}
\end{figure}

Two BCI calibration scenarios were considered~\cite{wu2016online}, as shown in Fig.~\ref{fig:settings}:
\begin{enumerate}
\item \emph{Offline UDA}, where the unlabeled test data from the target domain are accessible.
\item \emph{Online SDA}, where a small amount of labeled data from the target domain are accessible, but the target test data are inaccessible during training.
\end{enumerate}

Three cross-headset transfer tasks were studied: 1) BNCI2014001 $\rightarrow$ BNCI2014004 (only the left-hand and right-hand categories were used in BNCI2014001); 2) BNCI2015001 $\rightarrow$ BNCI2014002; and, 3) BNCI2014009 $\rightarrow$ BNCI2014008. Each task included offline and online calibration scenarios.

We assumed that the label spaces of the source and target domains are consistent. In online calibration, only one batch of labeled target data were accessible during training, to minimize the calibration effort as much as possible. For the MI paradigm, the classification accuracy was employed as the evaluation metric. For the P300 paradigm, since the datasets were highly class-imbalanced (non-target:target$\approx$5:1), the area under the curve (AUC) was utilized for evaluation.

For each group of transfer tasks, each target subject was treated as the target domain once, all algorithms were repeated five times with different random seeds, and the average performance of the five repeat was reported. All algorithms used EEGNet~\cite{Lawhern2018} as the backbone network, with batch size 32, learning rate $10^{-3}$, and the Adam optimizer in training. The temperature coefficient ${\tau}=2$ was used in SDDA. The trade-off parameters $\alpha$, $\beta$ and $\gamma$ were all set to 1.

All algorithms were implemented in PyTorch, and the source code is available on GitHub\footnote{https://github.com/Dingkun0817/SDDA}.

\subsection{Main Results}

We compared SDDA with nine existing deep learning transfer learning algorithms, including EEGNet~\cite{Lawhern2018}, DAN~\cite{long2015learning}, DANN~\cite{Ganin2016}, JAN~\cite{Long2017JAN}, CDAN~\cite{long2018conditional}, MDD~\cite{zhang2019bridging}, MCC~\cite{jin2020minimum}, SHOT~\cite{liang2020we}, and ISFDA~\cite{li2021imbalanced}. In online calibrations, we also included a traditional baseline, CSP-LDA (linear discriminant analysis)~\cite{blankertz2007optimizing} for MI, and xDAWN-LDA~\cite{rivet2009xdawn} for P300.

Tables~\ref{tab:MI_1}-\ref{tab:P300_1} show the results. Our proposed SDDA always achieved the best average performance, in both online SDA and offline UDA calibrations, for both MI and P300.

\begin{table*}[htpb]     \centering
\fontsize{9}{12}\selectfont %
       \caption{Classification accuracies  (\%) in BNCI2014001$\to$BNCI2014004 transfer. The best accuracies are marked in bold, and the second best by an underline.}  \label{tab:MI_1}
    \begin{tabular}{c|c|c|c|c|c|c|c|c|c|c|c}   \toprule
        Setting & Approach & S0 & S1 & S2 & S3 & S4 & S5 & S6 & S7 & S8 & Avg. \\
        \midrule
         \multirow{10}{*}{Offline Calibration}
         & EEGNet & \underline{66.53} & 55.62 & 57.67 & 84.92 & 74.54 & 68.70 & 67.95 & 75.84 & 70.58& 69.15$_{\pm0.70}$ \\
        & DAN & 65.67 & 55.85 & 57.17 & 86.27 & 74.73 & 69.97 & 70.44 & 75.92 & 70.56& 69.62$_{\pm0.51}$ \\
        ~ & DANN & 65.08 & 55.21 & 58.03 & 84.78 & 74.16 & 70.25 & 68.44 & \underline{76.71} & 72.53& 69.47$_{\pm0.62}$ \\
        ~ & JAN & 66.39 & 55.77 & 57.39 & 83.22 & 75.46 & 72.11 & 67.47 & 75.00 & 70.36& 69.24$_{\pm0.43}$ \\
        ~ & CDAN & 65.19 & \underline{56.62} & 58.36 & 85.84 & 75.16 & 73.28 & 69.53 & 75.08 & 71.11& 70.02$_{\pm0.33}$ \\
        ~ & MDD & 65.28 & 55.50 & \textbf{58.58} & 87.00 & 72.51 & 71.17 & 69.22 & 76.37 & 70.83& 69.61$_{\pm0.24}$\\
        ~ & MCC & 63.44 & 55.18 & 54.47 & \underline{91.95} & \underline{77.95} & \underline{74.33} & \underline{73.47} & 76.16 & 67.92& 70.54$_{\pm0.57}$ \\
        ~ & SHOT & 63.58 & 55.24 & 56.83 & 91.89 & 77.35 & 71.50 & 71.53 & 75.11 & \underline{73.72}& \underline{70.75}$_{\pm0.54}$ \\
        ~ & ISFDA & 64.75 & 56.06 & \underline{58.50} & 84.95 & 71.97 & 67.61 & 68.47 & 75.53 & 70.94& 68.75$_{\pm0.48}$ \\ \cline{2-12}
        ~ & SDDA (Ours) & \textbf{69.94} & \textbf{57.79} & 57.06 & \textbf{93.95} & \textbf{86.27} & \textbf{79.58} & \textbf{76.47} & \textbf{76.84} & \textbf{77.94} & \textbf{75.10}$_{\pm0.31}$ \\
        \midrule
        \multirow{11}{*}{Online Calibration}
        & CSP+LDA & 63.66 & \underline{56.17} & 54.94 & 88.42 & \underline{75.28} & \textbf{75.00} & 68.75 & \underline{77.89} & \underline{74.86} & 70.55\\
        ~ & EEGNet & 66.34 & 53.61 & 56.77 & 89.97 & 73.39 & 71.40 & 70.00 & 76.54 & 70.29 & 69.81$_{\pm0.52}$\\
        ~ & DAN & 66.48 & 53.92 & \underline{57.15} & 90.34 & 72.85 & 71.74 & 71.80 & 77.47 & 72.18 & 70.44$_{\pm0.23}$\\
        ~ & DANN & 65.29 & 55.32 & 55.81 & 89.83 & 74.83 & 70.81 & 67.09 & 77.23 & 72.04 & 69.82$_{\pm0.35}$\\
        ~ & JAN & 66.98 & 54.51 & 56.54 & 88.33 & 74.58 & 70.23 & 71.40 & 76.81 & 70.20 & 69.95$_{\pm0.39}$ \\
        ~ & CDAN & 66.80 & 54.63 & 56.89 & 89.83 & 75.09 & 71.42 & \textbf{72.91} & 76.48 & 72.06 & 70.68$_{\pm0.64}$ \\
        ~ & MDD & \underline{67.50} & 54.85 & 55.67 & 92.60 & \underline{75.28} & 71.34 & 70.96 & 77.01 & 71.19 & \underline{70.71}$_{\pm0.35}$ \\
        ~ & MCC & 67.09 & 55.09 & 55.99 & \underline{92.83} & 73.31 & 70.64 & 72.21 & 77.25 & 70.35 & 70.53$_{\pm0.41}$ \\
        ~ & SHOT & 65.09 & \underline{56.17} & 56.40 & 86.24 & 74.38 & 72.21 & 71.42 & 76.95 & 68.66 & 69.73$_{\pm0.60}$ \\
        ~ & ISFDA & 60.55 & 54.72 & \textbf{58.08} & 87.83 & 72.03 & 69.42 & 67.65 & 75.93 & 67.06 & 68.14$_{\pm0.30}$ \\ \cline{2-12}
        ~ & SDDA (Ours) & \textbf{70.73} & 56.02 & 57.09 & \textbf{93.96} & \textbf{78.25} & \underline{74.65} & \underline{72.53} & \textbf{79.45} & \textbf{75.44} & \textbf{73.12}$_{\pm0.34}$\\
        \bottomrule
    \end{tabular}
\end{table*}

\begin{table*}[htpb]
\addtolength{\tabcolsep}{-4pt}     \centering
\fontsize{9}{15}\selectfont %
       \caption{Classification accuracies  (\%) in BNCI2015001$\to$BNCI2014002 transfer. The best accuracies are marked in bold, and the second best by an underline.}  \label{tab:MI_2}
    \begin{tabular}{c|c|c|c|c|c|c|c|c|c|c|c|c|c|c|c|c}   \toprule
        Setting & Approach & S0 & S1 & S2 & S3 & S4 & S5 & S6 & S7 & S8 & S9 & S10 & S11 & S12 & S13& Avg. \\          \midrule
        \multirow{10}{*}{Offline Calibration}
        & EEGNet & 68.40 & 76.00 & 73.20 & 71.80 & 73.80 & 59.80 & 85.00 & 66.20 & 86.20 & 61.80 & 73.60 & 57.20 & 56.00 & 47.20 & 68.30$_{\pm1.07}$\\
        ~ & DAN & 69.20 & 77.40 & 68.40 & 66.80 & 76.20 & 60.00 & 85.80 & 67.20 & 85.00 & 59.80 & 79.00 & 58.80 & 56.60 & 47.20 & 68.39$_{\pm0.84}$\\
        ~ & DANN & 68.40 & 70.20 & 63.80 & 72.20 & 77.60 & 57.00 & 83.60 & \underline{68.40} & 84.40 & 63.80 & 76.20 & 57.80 & 57.00 & 49.20& 67.83$_{\pm1.08}$ \\
        ~ & JAN & \underline{72.80} & 76.60 & 76.60 & 70.20 & 78.40 & 61.40 & 86.40 & 65.20 & 82.20 & \underline{65.80} & 76.60 & 61.60 & 56.80 & 50.40& 70.07$_{\pm0.52}$ \\
        ~ & CDAN & 69.00 & 68.20 & 86.40 & 70.60 & 80.00 & 55.60 & 82.20 & 63.80 & 82.80 & 61.20 & 74.80 & \underline{62.00} & 55.80 & \textbf{53.80}& 69.01$_{\pm0.55}$ \\
        ~ & MDD & 71.40 & \underline{78.20} & 67.20 & \underline{73.60} & 74.60 & 59.00 & \underline{88.40} & 64.20 & 85.80 & 63.60 & 73.80 & 61.00 & 57.20 & 51.80& 69.27$_{\pm0.83}$ \\
        ~ & MCC & 71.40 & \underline{78.20} & \underline{96.60} & 69.80 & \underline{83.00} & \underline{62.00} & \textbf{89.80} & 62.20 & \underline{91.00} & 62.80 & \underline{80.60} & 61.80 & 55.40 & 49.60 & \underline{72.44}$_{\pm0.67}$ \\
        ~ & SHOT & 68.60 & \textbf{81.00} & 66.20 & 69.80 & 79.60 & 59.40 & 87.00 & 68.20 & 89.80 & 62.80 & 75.60 & 58.60 & \underline{60.80} & 51.00& 69.89$_{\pm0.80}$ \\
        ~ & ISFDA & 67.80 & 76.80 & 64.60 & 71.60 & 73.80 & 59.40 & 84.60 & 64.40 & 83.60 & 60.40 & 74.00 & 57.40 & 59.00 & \underline{52.00}& 67.81$_{\pm0.47}$ \\ \cline{2-17}
        ~ & SDDA (Ours) & \textbf{74.00} & 77.00 & \textbf{98.40} & \textbf{75.40} & \textbf{86.60} & \textbf{69.60} & 86.80 & \textbf{79.00} & \textbf{92.80} & \textbf{66.80} & \textbf{89.40} & \textbf{63.80} & \textbf{61.40} & 46.20 & \textbf{76.23}$_{\pm0.50}$\\
        \midrule
         \multirow{11}{*}{Online Calibration}
         & CSP+LDA & 58.82 & 72.06 & 91.18 & 64.71 & \textbf{77.94} & \underline{60.29} & \textbf{85.29} & \textbf{77.94} & \textbf{92.65} & 55.88 & 60.29 & \textbf{60.29} & 45.59 & 42.65 & 67.54 \\
        ~ & EEGNet & 69.71 & 75.29 & 91.47 & 66.47 & 71.47 & \textbf{61.47} & 81.47 & 63.53 & 84.41 & \underline{62.06} & 70.59 & 53.24 & 52.35 & 47.94& 67.96$_{\pm0.59}$ \\
        ~ & DAN & 67.06 & 74.41 & 90.29 & 65.29 & 73.82 & 58.24 & 81.76 & 63.53 & 87.06 & 60.29 & 72.35 & 57.35 & 54.71 & 51.76 & 68.42$_{\pm0.88}$\\
        ~ & DANN & \textbf{73.82} & 73.82 & 95.00 & 69.71 & 70.59 & 58.24 & 81.47 & 62.35 & \underline{90.88} & 59.71 & 68.53 & 53.24 & 50.88 & 50.00& 68.45$_{\pm0.65}$ \\
        ~ & JAN & 70.88 & \underline{79.12} & 80.59 & 73.53 & 71.18 & 50.59 & 83.82 & 64.41 & 85.88 & \textbf{63.53} & 72.94 & 54.12 & 52.06 & 51.18& 68.13$_{\pm0.82}$ \\
        ~ & CDAN & 65.29 & 71.76  & \textbf{95.59} & \underline{74.12} & 65.59 & 55.00 & 78.24 & 58.82 & 85.00 & 57.94 & 64.71 & 56.47 & 54.12 & 55.88 & 67.04$_{\pm1.05}$\\
        ~ & MDD & 70.29 & 77.06 & 90.88 & 72.06 & 71.47 & 55.00 & \textbf{85.29} & \underline{68.82} & 86.47 & 57.94 & \underline{72.65} & 52.06 & 47.06 & 49.71 & 68.34$_{\pm1.11}$\\
        ~ & MCC & 67.65 & \textbf{80.59} & 91.47 & 72.65 & 73.24 & 53.53 & 79.12 & 64.12 & 87.65 & 61.76 & 70.88 & 54.41 & 52.94 & 48.82 & \underline{68.49}$_{\pm1.07}$ \\
        ~ & SHOT & \underline{71.18} & 78.82 & 60.59 & 70.29 & 73.24 & 56.18 & 83.53 & 64.12 & 86.18 & 60.29 & 67.65 & \underline{59.12} & \underline{56.18} & \textbf{57.94}& 67.52$_{\pm0.61}$ \\
        ~ & ISFDA & 70.88 & 76.76 & 62.65 & \textbf{74.71} & 73.24 & 57.35 & 79.71 & 63.82 & 81.18 & 60.00 & 67.65 & 55.88 & \textbf{57.06} & \underline{57.65}& 67.04$_{\pm1.01}$ \\ \cline{2-17}
        ~ & SDDA (Ours) & 67.94 & 73.24 & \underline{94.12} & 72.94 & \underline{76.47} & 58.82 & \underline{84.12} & 68.24 & 87.65 & 61.18 & \textbf{82.06} & 57.65 & 53.53 & 55.59& \textbf{70.97}$_{\pm0.52}$ \\
    \bottomrule
    \end{tabular}
\end{table*}

\begin{table*}[htpb]     \centering
\fontsize{9}{12}\selectfont %
       \caption{Classification AUCs  (\%) in BNCI2014009$\to$BNCI2014008 transfer. The best AUCs are marked in bold, and the second best by an underline.}  \label{tab:P300_1}
    \begin{tabular}{c|c|c|c|c|c|c|c|c|c|c}   \toprule
        Setting & Approach & S0 & S1 & S2 & S3 & S4 & S5 & S6 & S7 & Avg. \\
        \midrule
         \multirow{10}{*}{Offline Calibration}
         & EEGNet & 74.45 & 66.55 & 79.23 & 67.46 & 68.48 & 69.78 & 68.68 & 77.05& 71.46$_{\pm0.23}$ \\
       ~ & DAN & 75.21 & 67.40 & 79.42 & 67.79 & 68.93 & 71.80 & 70.00 & 77.85 & 72.30$_{\pm0.39}$ \\
        ~ & DANN & 74.46 & 66.06 & 79.95 & 67.87 & 68.54 & 70.48 & 69.16 & 77.19 & 71.71$_{\pm0.32}$ \\
        ~ & JAN & 75.85 & 68.90 & 79.85 & 68.48 & 69.60 & 71.91 & 71.42 & 80.13 & 73.27$_{\pm0.18}$ \\
        ~ & CDAN & 76.04 & 69.41 & 80.43 & 68.53 & 70.65 & 73.74 & 72.40 & 81.53 & 74.09$_{\pm0.39}$ \\
        ~ & MDD & 74.93 & 66.34 & 79.69 & 67.58 & 69.15 & 71.07 & 69.17 & 76.29 & 71.78$_{\pm0.33}$ \\
        ~ & MCC & \underline{76.75} & \underline{69.56} & \underline{80.82} & 69.31 & \textbf{74.95} & \underline{74.59} & \underline{72.89} & \textbf{86.23} & \underline{75.64} $_{\pm0.19}$ \\
        ~ & SHOT & 74.92 & 66.71 & 79.53 & \underline{70.77} & 72.85 & 72.49 & 72.33 & 83.65 & 74.16$_{\pm0.61}$ \\
        ~ & ISFDA & 58.30 & 52.77 & 59.08 & 55.14 & 71.21 & 54.28 & 61.48 & 71.76& 60.50$_{\pm1.28}$ \\ \cline{2-11}
        ~ & SDDA (Ours) & \textbf{77.90} & \textbf{72.20} & \textbf{81.04} & \textbf{71.79} & \underline{73.84} & \textbf{77.20} & \textbf{74.65} & \underline{85.01} & \textbf{76.70}$_{\pm0.12}$ \\
        \midrule
        \multirow{11}{*}{Online Calibration}
        & xDAWN+LDA & 74.34 & 66.03 & 76.84 & 65.88 & 67.50 & 68.55 & 67.90 & 68.00 & 69.38 \\
        ~ & EEGNet & \underline{77.75} & 74.00 & 81.78 & 71.94 & 71.88 & 77.94 & 80.47 & 88.41 & 78.02$_{\pm0.23}$\\
        ~ & DAN & 76.81 & 74.17 & 81.67 & 72.10 & 72.92 & 78.11 & 80.88 & \underline{88.68} & 78.17$_{\pm0.34}$ \\
        ~ & DANN & 76.94 & \underline{74.48} & 81.10 & \underline{72.30} & 73.37 & 78.51 & 80.29 & 87.56 & 78.07$_{\pm0.56}$ \\
        ~ & JAN & 77.62 & \textbf{74.58} & 81.99 & \textbf{72.84} & 72.48 & \underline{79.82} & \underline{81.61} & 87.87 & \underline{78.60}$_{\pm0.29}$ \\
        ~ & CDAN & 76.94 & 73.09 & \underline{82.15} & 72.31 & 72.68 & 78.59 & 80.58 & 87.47& 77.98$_{\pm0.33}$ \\
        ~ & MDD & 77.53 & 74.33 & 81.72 & 71.74 & 72.79 & 79.55 & 80.47 & 88.38 & 78.31$_{\pm0.24}$ \\
        ~ & MCC & 77.70 & 74.35 & 81.63 & 71.58 & 72.62 & 77.67 & 80.90 & \textbf{88.85} & 78.16$_{\pm0.22}$ \\
        ~ & SHOT & 76.19 & 67.92 & 79.86 & 68.76 & 70.55 & 70.82 & 72.62 & 79.19 & 73.24$_{\pm0.56}$ \\
        ~ & ISFDA & 77.53 & 69.43 & 81.86 & 71.76 & 73.54 & 70.29 & 72.05 & 82.51 & 74.87$_{\pm0.87}$ \\ \cline{2-11}
        ~ & SDDA (Ours) & \textbf{79.87} & 73.27 & \textbf{83.17} & 67.15 & \textbf{77.99} & \textbf{84.28} & \textbf{82.75} & 87.25& \textbf{79.47}$_{\pm0.37}$ \\
        \bottomrule
    \end{tabular}
\end{table*}

\subsection{Ablation Studies}

Ablation studies were performed on six variants of SDDA to evaluate the contributions of each individual components:
\begin{enumerate}
\item CE, which uses only the source domain cross-entropy loss.
\item CE+SD, which adds SD to CE.
\item CE+MA, which adds MA to CE.
\item CE+CL, which adds CL to CE.
\item CE+MA+CL, which adds MA and CL to CE.
\item SDDA, which is CE+SD+MA+CL.
\end{enumerate}

\begin{figure*}[htpb]
	\centerline{\includegraphics[width=1\linewidth]{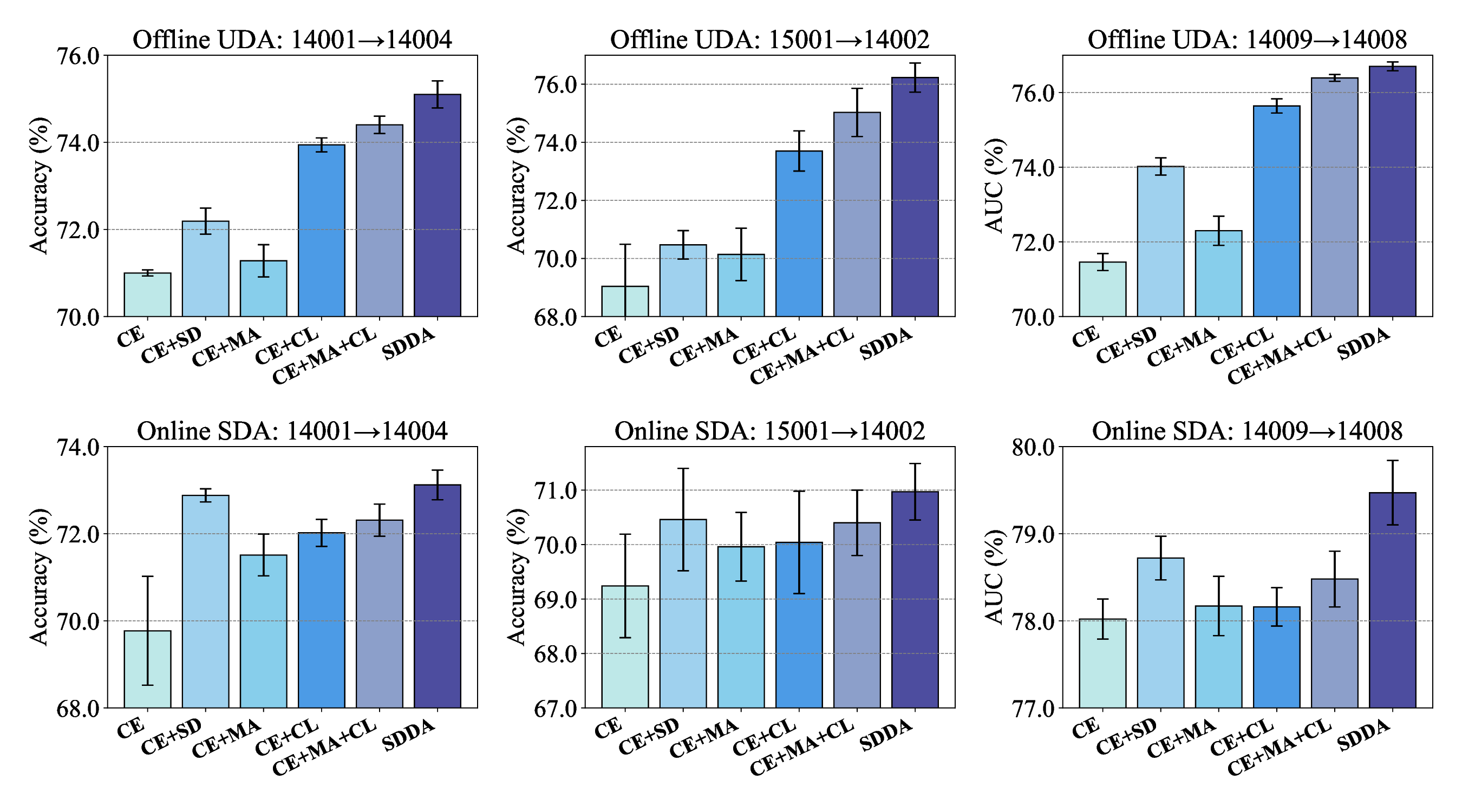}}
	\caption{Ablation study results.} 	\label{fig:ablation}
	\end{figure*}

As shown in Fig.~\ref{fig:ablation}, in both BCI paradigms and both calibration scenarios, adding SD, MA or CL to CE always improved the performance of CE, and adding MA and CL together always outperformed adding MA or CL alone. SDDA, which includes all four components (CE, SD, MA and CL), always achieved the best performance.

\subsection{Effectiveness of EA}

\begin{figure*}[htbp]\centering
\subfigure[]{\includegraphics[width=0.46\linewidth,clip]{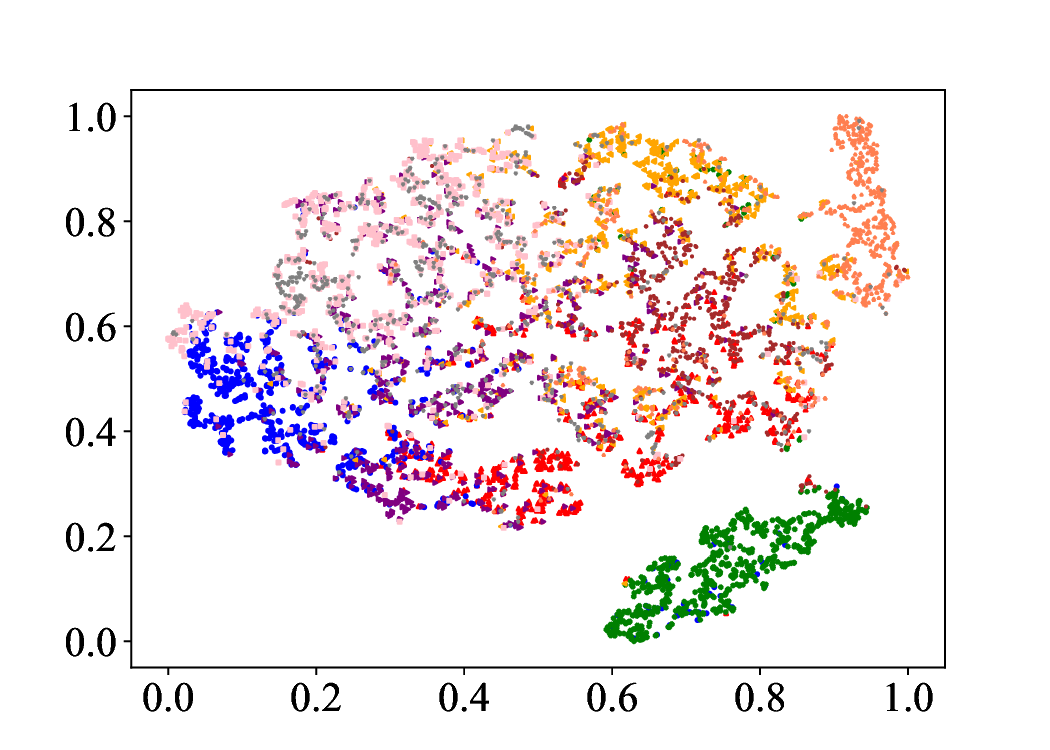}}
\subfigure[]{\includegraphics[width=0.52\linewidth,clip]{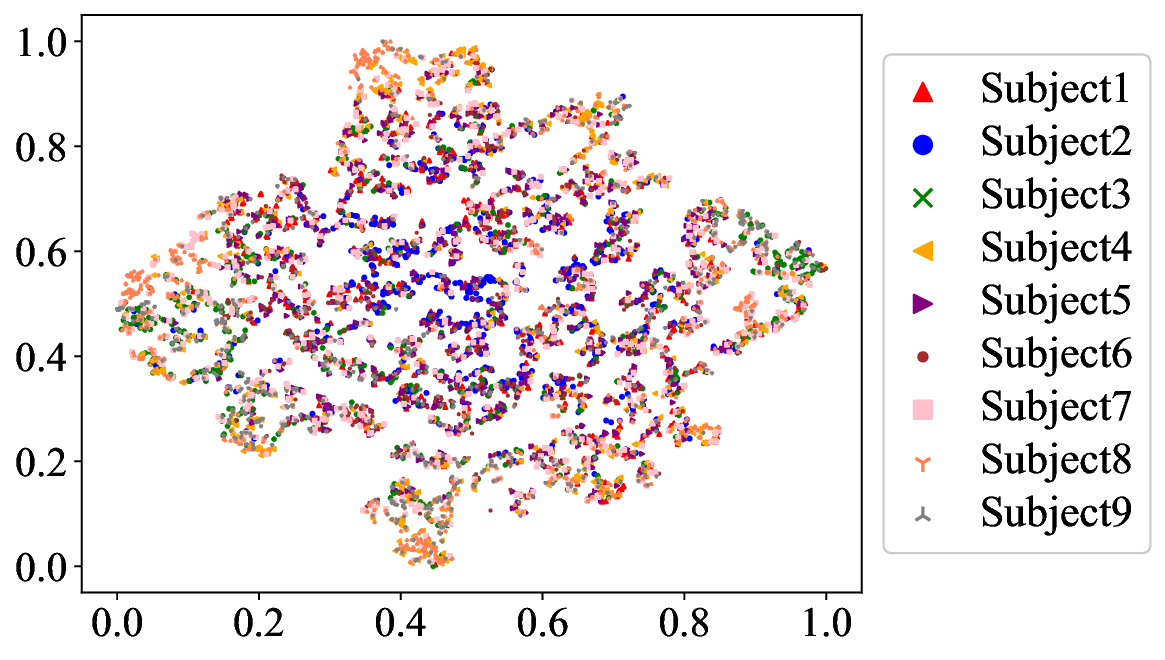}}
\caption{$t$-SNE visualization of the data in BNCI2014004. (a) Before EA; (b) After EA. Different colors represent trials from different subjects.} \label{fig:sEA}
\end{figure*}

$t$-distributed Stochastic Neighbor Embedding ($t$-SNE)~\cite{Maaten2008}, a widely used dimensionality reduction technique, was used to illustrate the effectiveness of data alignment. Fig.~\ref{fig:sEA} shows the results. Clearly, after EA, EEG trials from different subjects became more consistent, facilitating transfer.

\subsection{Comparison with Homogeneous Transfer}

To demonstrate the necessity of making use of the extra channels in the source domain, we compared SDDA with homogeneous transfer methods that use only the common subset of channels of the two domains. Table~\ref{tab:004_upper} shows the results. SDDA consistently outperformed all homogeneous transfer learning algorithms, underscoring the importance of leveraging additional channel information from the source dataset.

\begin{table*}[htpb]     \centering
\fontsize{9}{12}\selectfont %
       \caption{Classification accuracies (\%) of homogeneous and heterogeneous transfers on BNCI2014004. The best accuracies are marked in bold, and the second best by underline.}  \label{tab:004_upper}
    \begin{tabular}{c|c|c|c|c|c|c|c|c|c|c|c}   \toprule
        Setting & Approach & S0 & S1 & S2 & S3 & S4 & S5 & S6 & S7 & S8 & Avg. \\
        \midrule
         \multirow{10}{*}{Homogeneous Transfer}
         & EEGNet & 70.03 & 58.09 & 58.00 & 89.60 & 73.51 & 74.92 & 71.19 & 79.45 & 77.64& 72.49$_{\pm0.60}$ \\
        & DAN & \underline{70.25} & 58.38 & 58.11 & 90.84 & 75.08 & 75.33 & 71.75 & \underline{79.97} & 78.33 & 73.12$_{\pm0.67}$ \\
        ~ & DANN & 68.83 & \textbf{59.41} & \textbf{59.00} & 87.60 & 74.95 & 76.72 & 70.06 & \textbf{80.79} & 78.97 & 72.93$_{\pm0.60}$ \\
        ~ & JAN & \textbf{70.97} & 57.77 & 57.75 & 88.95 & 75.68 & 77.64 & 72.14 & 78.74 & 78.08 & 73.08$_{\pm0.57}$ \\
        ~ & CDAN & 69.58 & 57.85 & 57.83 & 87.89 & 78.03 & 78.17 & \underline{73.00} & 79.32 & \underline{79.56} & 73.47$_{\pm0.37}$ \\
        ~ & MDD & 69.64 & \underline{58.65} & 57.33 & 88.97 & 74.16 & 78.56 & 71.92 & 79.87 & 79.25 & 73.15$_{\pm0.33}$\\
        ~ & MCC & 68.17 & 56.94 & \underline{58.92} & 90.16 & \underline{79.43} & 78.11 & 73.69 & 76.74 & \textbf{80.08}& \underline{73.58}$_{\pm0.64}$ \\
        ~ & SHOT & 69.61 & 58.35 & 57.78 & \textbf{94.46} & 75.68 & \underline{78.81} & 71.83 & 77.47 & 75.86 & 73.32$_{\pm1.24}$ \\
        ~ & ISFDA & 64.58 & 57.91 & 57.53 & 90.92 & 75.03 & 76.22 & 72.00 & 78.37 & 76.69& 72.14$_{\pm0.58}$ \\
        \midrule
        \multirow{1}{*}{Heterogeneous Transfer}
        & SDDA (Ours) & 69.94 & 57.79 & 57.06 & \underline{93.95} & \textbf{86.27} & \textbf{79.58} & \textbf{76.47} & 76.84 & 77.94 & \textbf{75.10}$_{\pm0.31}$\\
        \bottomrule
    \end{tabular}
\end{table*}

\section{Conclusions} \label{sect:conclusion}

This paper has proposed an SDDA algorithm for heterogeneous cross-headset transfer for BCI calibration. Existing transfer learning methods typically use only the common channels of the source and target domains, resulting in the loss of spatial information and suboptimal performance. SDDA uses first spatial distillation to make use of the full set of channels, and then input/feature/output space distribution alignments to cope with the significant differences between the source and target domains. To our knowledge, this is the first work to introduce knowledge distillation for cross-headset transfers. Extensive experiments on six EEG datasets from two BCI paradigms demonstrated that SDDA achieved superior performance in both offline unsupervised and online supervised domain adaptation scenarios, consistently outperforming 10 classical and state-of-the-art transfer learning algorithms.

\bibliographystyle{IEEEtran} \bibliography{sdda}

\end{document}